\documentclass[letterpaper, 10 pt, conference]{ieeeconf}  

\IEEEoverridecommandlockouts                              

\title{\LARGE \bf
Object Scan Context: Object-centric Spatial Descriptor for Place Recognition within 3D Point Cloud Map
}

\author{Haodong Yuan$^{1,2}$, Yudong Zhang$^{2}$, Shengyin Fan$^{2}$, Xue Li$^{2}$ and Jian Wang$^{1}$
\thanks{$^{1}$Yuan is the postgraduate student of School of Transportation Science and Engineering, Beihang University, China.
Wang, the tutor of Yuan, is with the school of Transportation Science and Engineering, Beihang University, China.
        {\tt\small \{yuanhd17\}, \{07248\}@buaa.edu.cn}}%
\thanks{$^{2}$Yuan, Zhang, Fan ans Li is with Faculty of Beijing Yihang Yuanzhi Technology Co., Ltd,
        {\tt\small \{zhangyudong@126.com\}, \{fanshengyin@126.com\}, \{lixue@yihang.ai\}}}%
}

\usepackage{amsmath}
\usepackage{graphicx}
\usepackage{float}
\usepackage{subfigure}
\usepackage{caption}
\usepackage{ulem}
\usepackage{booktabs}
\usepackage[flushleft]{threeparttable}
\usepackage{multirow}
\usepackage{threeparttable}
\begin{document}

\maketitle
\thispagestyle{empty}
\pagestyle{empty}

\begin{abstract}
The integration of a SLAM algorithm with place recognition technology empowers it with the ability to mitigate accumulated errors and to relocalize itself. However, existing methods for point cloud-based place recognition predominantly rely on the matching of descriptors, which are mostly lidar-centric. These methods suffer from two major drawbacks: first, they cannot perform place recognition when the distance between two point clouds is significant, and second, they can only calculate the rotation angle without considering the offset in the X and Y directions. To overcome these limitations, we propose a novel local descriptor that is constructed around the Main Object (\ref{Contruct Descriptors}). By using a geometric method (Section \ref{Estimate Similarity}), we can accurately calculate the relative pose. We have provided a theoretical analysis to demonstrate that this method can overcome the aforementioned limitations. Furthermore, we conducted extensive experiments on KITTI Odometry and KITTI360, which indicate that our proposed method has significant advantages over state-of-the-art methods.

\end{abstract}

\section{INTRODUCTION} \label{INTRODUCTION}

Intelligent vehicles have become a major research focus for enterprises, universities, and research institutions worldwide. Precise self-localization is crucial for the safe driving of intelligent vehicles. Several established positioning methods utilizing diverse sensors such as lidar, camera, imu, and GPS have been developed. Notably, Simultaneous Localization and Mapping (SLAM) has emerged as a favored technique for self-localization owing to its convenience, robustness, and map reusability.

A mature SLAM algorithm consists of three main components: the Front End, which estimates vehicle motion between frames; the Back End, which optimizes vehicle states and the global map; and Loop Closure, which helps eliminate cumulative errors and establish a globally consistent map. Additionally, to address sensor instability, a relocalization module is often included, which assists the vehicle in quickly determining its position after sensor wake-up.

\begin{figure}[thpb]
        \centering
        \includegraphics[width=0.45\textwidth]{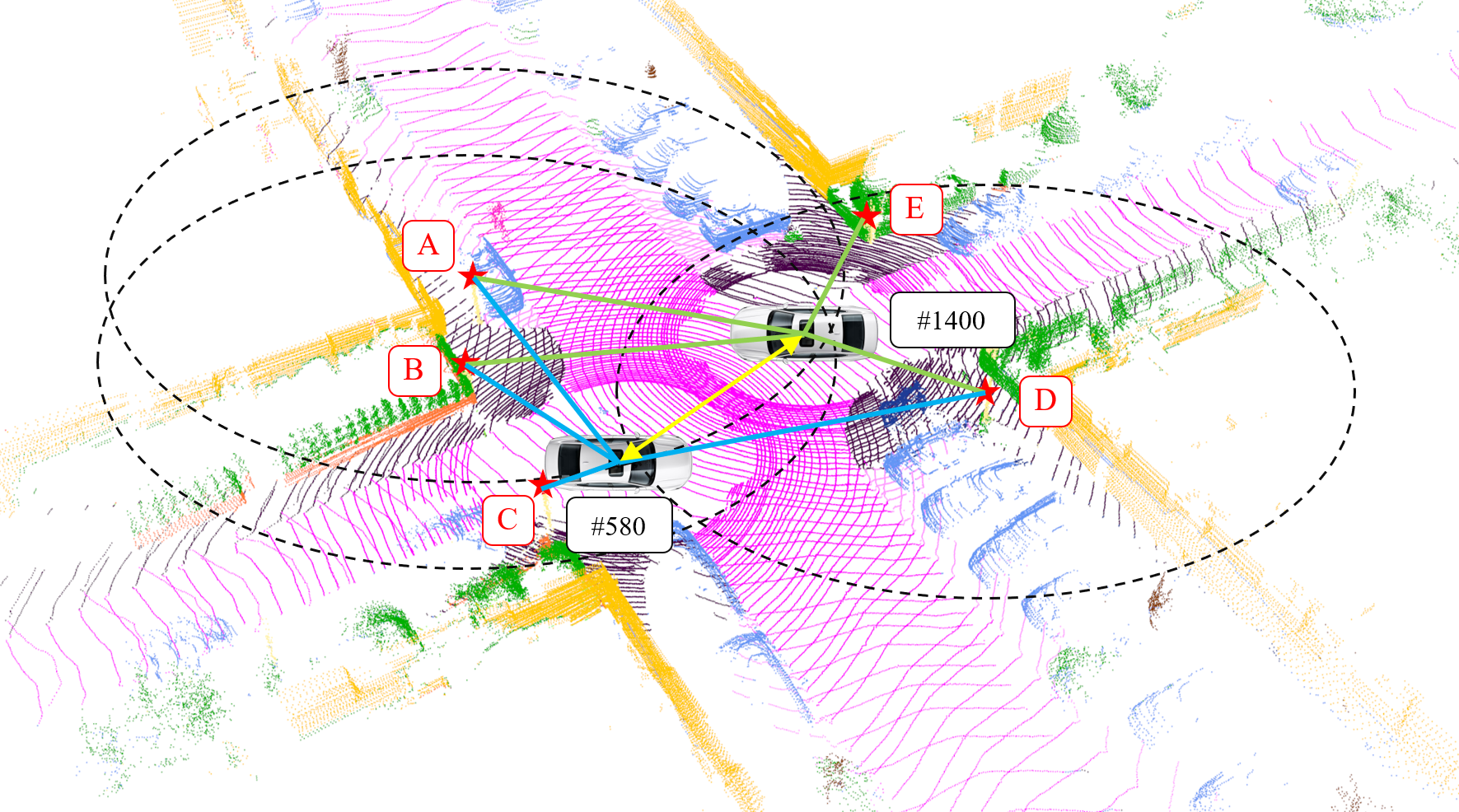}
        
        \caption{An example of place recognition using Object Scan Context. It is \#580 and \#1400 frame of KITTI sequence 00, which form a contactless loop. After semantic segmentation network and clustering processing, \#580 frame contains four Main Objects (A, B, C, D), and \#1400 frame contains another four (A, B, D, E). It only needs to match any repeated OSC (A, B, D) in the two frames to perform place recognition and calculate relative pose $(\Delta x, \Delta y, \Delta \theta)$.}
        
        \label{F1}
\end{figure}

SLAM, as a widely adopted technique, is susceptible to cumulative errors resulting from sensor noise and imprecise odometry measurements. Addressing this challenge entails employing high-precision sensors or integrating data from multiple sensor modalities. For instance, FAST-LIO \cite{FAST-LIO} leverages a filtering framework to fuse lidar and inertial data, while LIO-SAM \cite{LIO-SAM} employs factor graphs, and Zhu \cite{ZHU1} utilizes pose graphs for integrating GPS/NRTK/IMU/Vision data to achieve local and global positioning. Loop closure, a commonly used approach, helps mitigate cumulative errors. However, due to significant pose differences between frames, direct application of point cloud registration algorithms for loop closure detection is impractical. Thus, enhancing the accuracy and recall rate of loop closure detection and accurately estimating relative pose are key research objectives addressed in this paper.

Loop closure detection and relocalization encompass various techniques, including Euclidean distance search, descriptor-based place recognition, and sensor data registration. Among these, the main research focus of this paper lies in the domain of descriptor-based place recognition methods. Similar to the classification of SLAM, place recognition techniques can be categorized into two main approaches based on the type of sensor employed: image-based and point cloud-based, the latter being the focus of this paper. Point cloud-based methods typically involve transforming each frame of the point cloud into a descriptor, and matching these descriptors enables the computation of a similarity metric. Moreover, advanced algorithms can estimate the approximate relative pose between point clouds, thereby enhancing localization accuracy by leveraging the iterative closest point (ICP) algorithm \cite{ICP}.

Scan Context algorithms are typical examples of above processes, including Scan Context (SC) \cite{SC}, Intensity Scan Context (ISC) \cite{ISC}, Semantic Scan Context (SSC) \cite{SSC}, Scan Context++ (SC++) \cite{SC++}, and so on. These algorithms create a descriptor, known as the Scan Context, for each point cloud frame centered on the lidar. By matching descriptors through rotation, similarity and rotation angle can be calculated. Additionally, SSC includes a two-step semantic ICP algorithm, which can estimate the 3D pose. Although these algorithms have shown promising loop-closure detection performance on numerous datasets, they also possess significant and critical limitations.

\begin{itemize}

\item Scan Context is a lidar-centric descriptor, meaning that it only captures the surroundings from the current observation position. As a result, if the distance between two vehicle poses is not sufficiently small, the corresponding descriptors will differ significantly, leading to missed loop closures or unsuccessful relocalization attempts, which is called contactless cases in this paper.

\item The rotation matching method used in Scan Context algorithms can only estimate the yaw angle of the relative pose, and providing it as the initial pose for ICP may lead to convergence to a local optimum. Although recent methods such as Semantic Scan Context (SSC), OverlapNet(ON), and Scan Context++ (SC++) can compute the 3D pose, their performance in contactless place recognition is often hindered by error-prone calculations.

\end{itemize}

\begin{figure*}[thpb]
        \centering
        \includegraphics[width=1.0\textwidth]{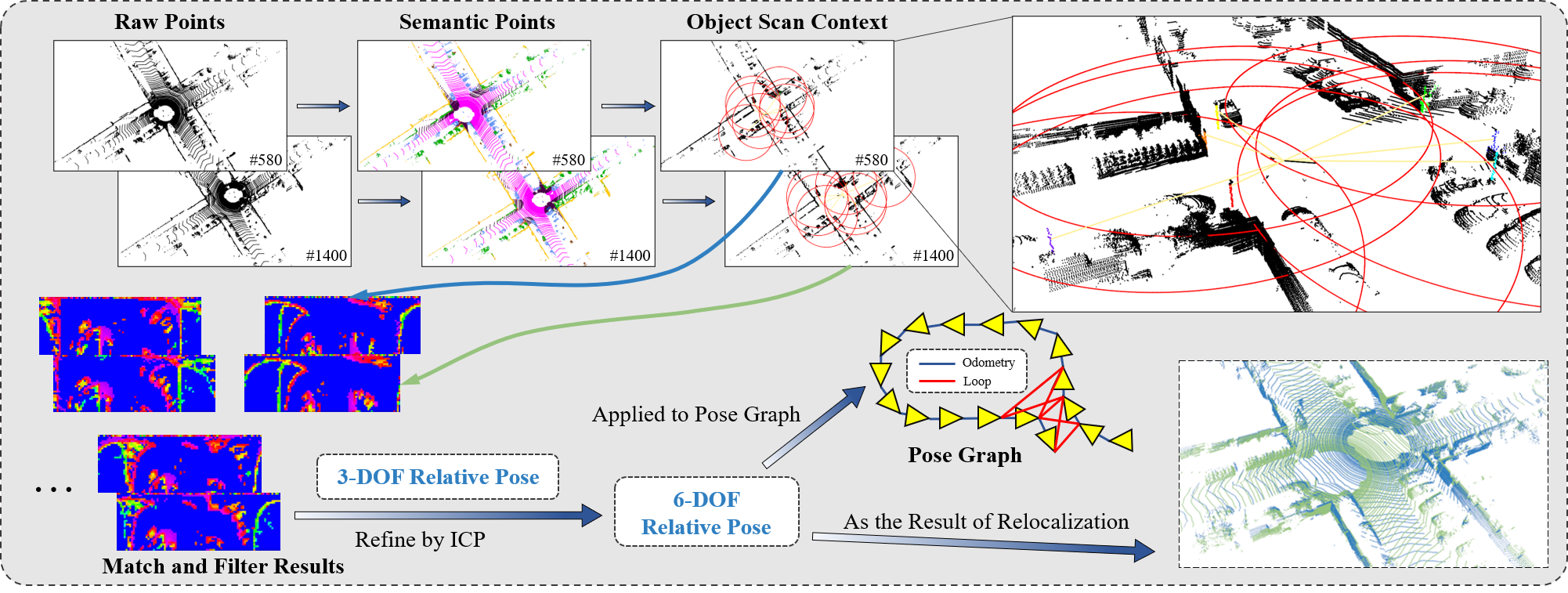}
        
        \caption{The architecture of our approach. The whole framework consists of building Object Scan Context and calculating similarity score and relative pose between frames. First, we perform semantic segmentation and Euclidean clustering on raw points. After that, Main Objects are achieved (the colored points in the enlarged image are Main Objects, and the red circles indicate the scopes of Object Scan Contexts). Second, an accurate relative pose will appear by matching descriptors. It is worth mentioning that the pose is very accurate so that it can be directly applied to Pose Graph or act as a result of relocalization.}
        
        \label{F2}
\end{figure*}

In this paper, a novel local descriptor called Object Scan Context (OSC) is proposed, which is designed to concentrate on the surrounding environment of certain prominent static objects referred to as the Main Objects, rather than the environment in the proximity of the vehicle. The selection of main objects will be introduced in the Section \ref{Get Main Objects}.

The unique improvement of constructing descriptors based on main objects makes the descriptor no longer dependent on the vehicle position, and can calculate accurate relative poses based on relative observations (strict geometric proof in \ref{Relative Pose}), which is difficult for SC series algorithms. However, this operation will change the descriptor from global to local, and a frame of point cloud will contain multiple descriptors, which will increase the complexity of the system. Therefore, we propose a series of algorithms to improve efficiency, including an accelerated search method based on RingKey (\ref{Get Candidate KeyFrames}), a bubble filtering algorithm to select the most robust matching results (\ref{Bubble Filter}), and a 3D pose filtering algorithm (\ref{Pose Filter}). Fig. \ref{F1} presents a demonstration of our results. The main contributions are summarized as follows:


\begin{itemize}

\item We propose a novel semantic-enhanced local descriptor that is independent of the lidar's position for point cloud-based place recognition, which focuses on the environment around Main Objects rather than the environment around the lidar itself.
\item We conducted a theoretical analysis of the process of creating and matching Object Scan Contexts. Our method can address two major drawbacks of other Scan Context algorithms, and the resulting relative pose can be utilized as an initial value for point cloud registration.
\item Exhaustive experiments conducted on the KITTI Odometry and KITTI360 datasets demonstrate that our proposed method achieves state-of-the-art performance in both place recognition and pose estimation.

\end{itemize}

\section{RELATED WORK} \label{RELATED WORK}

In this article, we will only discuss place recognition methods based on 3D point clouds. Traditional methods extract some geometric features from point clouds, while some methods take advantage of neural networks' superior feature extraction capabilities to obtain advanced features from point clouds, such as semantics. Some methods use neural networks to perform descriptor matching, hoping to obtain deeper information associations. According to whether to use neural network and the way neural networks are used, we divide place recognition algorithms into three categories: pure geometric methods, learning-based descriptor construction methods, and learning-based descriptor matching methods.

\textbf{Pure geometric methods: }Scan Context \cite{SC} projects the 3D point cloud onto the X-Y plane and creates a global descriptor capable of performing rotation matching by dividing it into Rings and Sectors. Scan Context++ \cite{SC++} builds two descriptors, namely Polar Context and Cart Context, for a 3D point cloud, providing Scan Context++ with invariance in both rotational and lateral directions, with the lateral invariance specifically designed to better handle contactless place recognition. Intensity Scan Context \cite{ISC} is established using the same approach as Scan Context, replacing the maximum height with the intensity as the feature of the grid. M2DP \cite{M2DP} is built from the projection of the 3D point cloud onto multiple 2D planes, and Singular Value Decomposition (SVD) is then used to reduce the dimensions of the final descriptor. ISHOT \cite{ISHOT} proposes a local descriptor combining geometric and texture information from LiDAR intensity, as well as a probabilistic keypoint voting place recognition method. LiDAR Iris \cite{IRIS} proposes a global descriptor for LiDAR point clouds based on a binary signature image obtained through Gabor filtering and thresholding operations. Lin \textit{et al.} \cite{FAST} compare the 2D histograms of plane and line features of two keyframes to calculate their similarity and integrate it into the loop detection module of the SLAM system for experimentation. The above methods are all based on the geometric features of point clouds to build descriptors, which can achieve good results in some scenarios.

\textbf{Learning-based descriptor construction methods: }Semantic Scan Context \cite{SSC} uses semantic information to replace the height of the point to construct the Scan Context, and cooperates with its proposed two-step ICP, which can improve the descriptor matching accuracy and generate a relative pose for point cloud registration. SA-LOAM \cite{SA-LOAM} uses a semantic segmentation network to convert the original point cloud into a semantic point cloud, then generates a semantic graph to express the relative position of various semantics in the point cloud, and calculates the similarity of the pair of graphs by the network. SA-LOAM starts to pay attention to the invariance of the relative position between objects, but its descriptor is still closely dependent on the observation position, and it does not actually extract the object, so it can only alleviate the two major flaws rather than completely solve them. PointNetVLAD \cite{POINTNET} applies metric learning to generate a discriminative and compact global descriptor from an unordered input 3D point cloud and proposes a novel loss function that make descriptors more discriminative and generalizable. SeqSphereVLAD \cite{SEQSPHEREVLAD} uses spherical projection and a neural network with four Spherical Convolution Layers, four WAG Pooling Layers and a Flatten Layer to generate an orientation-invariant place descriptor. GOSMatch \cite{GOSMATCH} proposes a semantic-based graph descriptor, which pays attention to the transformation relationship between semantics and semantics in the scene, and gives a 6-DOF initial pose estimation.

\textbf{Learning-based descriptor matching methods: }OverlapNet \cite{OVERLAPNET} creates a multi-head neural network to calculate the yaw angle and similarity between any two point clouds, which is called the overlap in original article. OverlapNetTransformer \cite{OVERLAPNETTRANSFORMER} adds a transformer attention mechanism to OverlapNet to improve its robustness and operational efficiency. SGPR \cite{SGPR} uses a semantic graph to represent the point cloud, and then learns the similarity between two semantic graphs. Furthermore, the authors prove the feasibility and robustness of their work and apply it to SA-LOAM \cite{SA-LOAM}. LCDNet \cite{LCDNET} combines a shared feature extractor, discriminative global descriptors, and a differentiable relative pose head based on unbalanced optimal transport thoery. In addition, its integration with LIO-SAM \cite{LIO-SAM} forms a complete SLAM system capable of loop detection in the presence of strong drift.

\section{METHODOLOGY} \label{METHODOLOGY}
In this section, we formally present our method. Different from the existing Scan Context methods that build a local coordinate system with lidar as the center, we encode the local environment with Main Objects as the center. We present the encoding method and similarity calculation method for the descriptors. We provide a formula for computing the 3-DOF relative pose based on coordinate transformation, and propose a matching strategy to ensure the efficiency and robustness of the algorithm. The framework is shown in Fig. \ref{F2}.

\begin{figure}[thpb]
        \centering
        \includegraphics[width=0.45\textwidth]{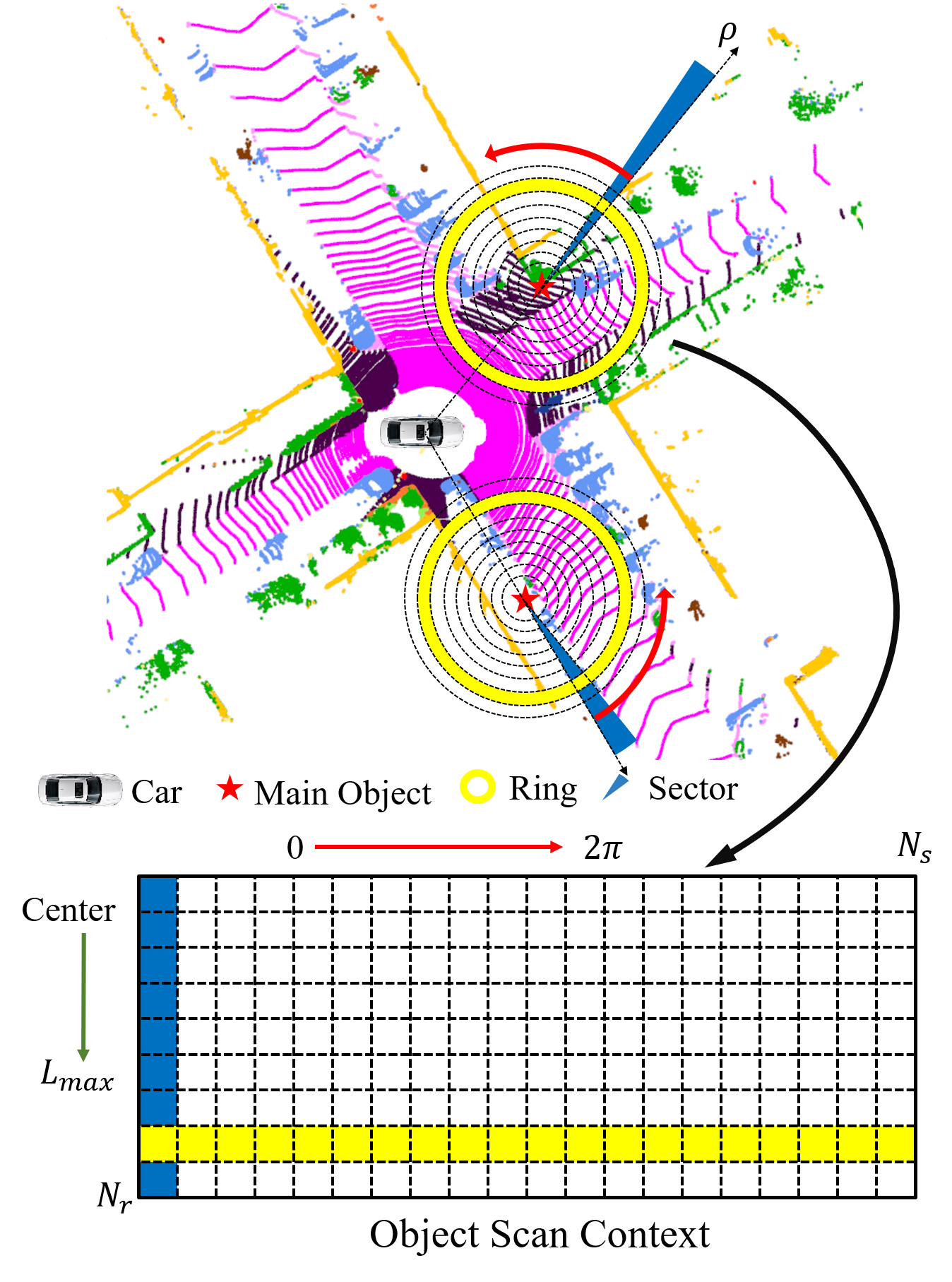}
        
        \caption{An example of generating Object Scan Context. We build OSC with Main Object $(red \ star)$ as the center. As shown in the figure above, OSC is a matrix with shape $(N_r \times N_s)$, where Sector is the column of the matrix and Ring is the row. Calculate the average of each column of OSC to get a row vector called SectorKey, and calculate the average of each row to get a column vector called RingKey.}
        
        \label{F3}
\end{figure}

\subsection{Contruct Descriptors} \label{Contruct Descriptors}
\textbf{Get Main Objects.} \label{Get Main Objects} When humans explore a new environment, they typically rely on the most prominent objects to orient themselves and determine their position. Motivated by this observation, we need to select one or several types of semantics to pay attention to (which we refer to as Main Objects) - usually static, immovable, and widely distributed objects such as street lights on urban roads. However, we also consider the surrounding environment to ensure sufficient diversity. For instance, tree trunks are usually surrounded by grass, making them unsuitable as Main Objects. The specific operational procedures are as follows.

For a frame of raw point cloud, we first infer the semantics of each point using a semantic segmentation network, and then filter out point clouds with the target semantics. Next, we perform Euclidean clustering and calculate the average coordinate of each point cloud cluster as the location of the Main Object.

\textbf{Build Object Scan Context.} \label{Build Object Scan Context} After obtaining the coordinates of all Main Objects, we encode a descriptor for each Main Object to describe its local environment.


In the X-Y plane, we consider the Main Object as the pole, and take the direction from the lidar to the Main Object as the polar diameter to establish a polar coordinate system (counterclockwise is set to be positive). Then, the circular area within a specific distance is divided into $N_r$ equal-width rings, known as Rings, and divided into $N_s$ equal-angle sectors, known as Sectors, resulting in $N_r \times N_s$ grids. After projecting the 3D point cloud onto the X-Y plane, we compute the average height of the points in each grid, which serves as the feature of the grid. Each Ring is considered as a row and each Sector as a column, generating a matrix of $N_r \times N_s$ called the Object Scan Context, as illustrated in Fig. \ref{F3}. The average value of each row of this matrix is computed to obtain a column vector known as RingKey, and the average value of each column is computed to obtain a row vector called SectorKey. Both will be used to enhance the system's efficiency in the future.

\subsection{Estimate Similarity} \label{Estimate Similarity}
\textbf{Get Candidate KeyFrames.} \label{Get Candidate KeyFrames} The above encoding method reveals that the descriptor of the same Main Object undergoes a horizontal translation (defined as dividing the matrix into two blocks from left to right and exchanging their positions) with varying positions of the lidar, but the RingKey is theoretically consistent. Therefore, to improve the efficiency of candidate search, we use RingKey as nodes in the KD-tree. 

Furthermore, in order to fully leverage all potential similar keyframes, we select multiple candidate descriptors from the KD-tree and consider all keyframes corresponding to them as candidate keyframes for matching. Finally, we obtain a set of candidate keyframes that are similar to the current frame.



\textbf{Calculate Similarity Score.} \label{Calculate Similarity Score} After obtaining candidate frame pairs, we need to determine if two frames are from the same location by matching the descriptors of these keyframes one by one. We use cosine distance between sectors to measure the similarity of two sectors, and average similarity of all sector pairs to measure the similarity of two Object Scan Contexts. The following section describes the process of matching two descriptors.

We shift the Object Scan Context of the current keyframe to the left by taking the first column of the matrix and placing it in the last column in turn. We need to calculate each similarity under each shift simultaneously. Usually, one similarity is significantly greater than the other values, and the corresponding offset will be used to calculate the yaw angle of the relative pose in the future. To improve the efficiency of this process, we use SectorKey to quickly determine the approximate offset, and then only need to calculate the similarity within an offset window instead of the entire range from 0 to Ns. The approximate offset $shift$ is calculated as follows:
\begin{equation}
        shift^*=\mathop{\arg\min}_{shift \in [0, N_s)}||S^q-S^c_{shift}||_2
\end{equation}
where $S^q$ represents the SectorKey of the candidate Object Scan Context, and $S^c_{shift}$ represents the SectorKey of the current Object Scan Context shifted to the left by $shift$ units, and $||\cdot||_2$ represents the 2-norm of a vector.

We use a search window with $shift^*$ as the center and $2k$ as the width to replace the entire range of $[0, N_s)$, which can greatly reduce the amount of calculation. Using $I^q$ and $I^c_n$ to indicate the candidate and the current Object Scan Context, where the subscript $n$ is used to indicate that the descriptor is shifted to the left by $n$ units. The following parameter optimization equation can be constructed to obtain the precise offset $n^*$ and the similarity between the descriptor pair:
\begin{equation}
        n^*=\mathop{\arg\min}_{n \in [shift^*-k, shift^*+k]}d(I^q, I^c_n)
\end{equation}
\begin{equation}
        similarity=1-d(I^q, I^c_{n^*})
\end{equation}
where $d$ represent an abstract distance between two descriptors, which can be derived from the cosine distance of the vector:
\begin{equation}
        d(I^q,I^c_n)={1 \over N_s}\sum^{N_s}_{j=1}(1-{c^q_j \cdot c^c_{jn} \over ||c^q_j|| \cdot ||c^c_{jn}||})
\end{equation}
where ${c^q_j \cdot c^c_{jn} \over ||c^q_j|| \cdot ||c^c_{jn}||}$ indicates the cosine distance between the $j$th Sector of the candidate descriptor and the $j$th Sector of the current descriptor after performing the precise offset $n^*$.

\begin{figure}[thpb]
        \centering
        \includegraphics[width=0.45\textwidth]{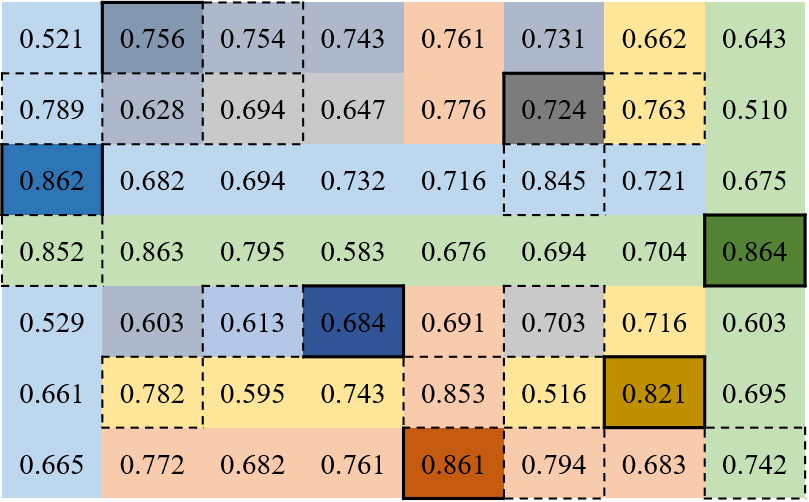}
        
        \caption{An example of the bubble filter. The solid box indicates the selected maximum value, while the dashed box indicates the selected second maximum value.}
        
        \label{F4}
\end{figure}

\textbf{Bubble Filter.} \label{Bubble Filter} Assuming that the candidate frame and the current frame contain $m$ and $n$ Object Scan Contexts respectively. After the previous two steps, $m \times n$ similarities are achieved. As shown in Fig. \ref{F4}, the matrix of $m$ rows and $n$ columns represents $m \times n$ matching results.

First, we select the maximum similarity and the second maximum similarity in the same row and column. The reason for selecting the second maximum value is to avoid missing matches. Then, remove the row and column corresponding to the selected maximum similarity from the matrix. Repeat this process until it can no longer be performed. At the end of the these processes, we obtain $min(m, n)$ maximum values and some second maximum values. In the next subsection, we will calculate the relative poses corresponding to these values.

\begin{figure}[thpb]
        \centering
        \includegraphics[width=0.45\textwidth]{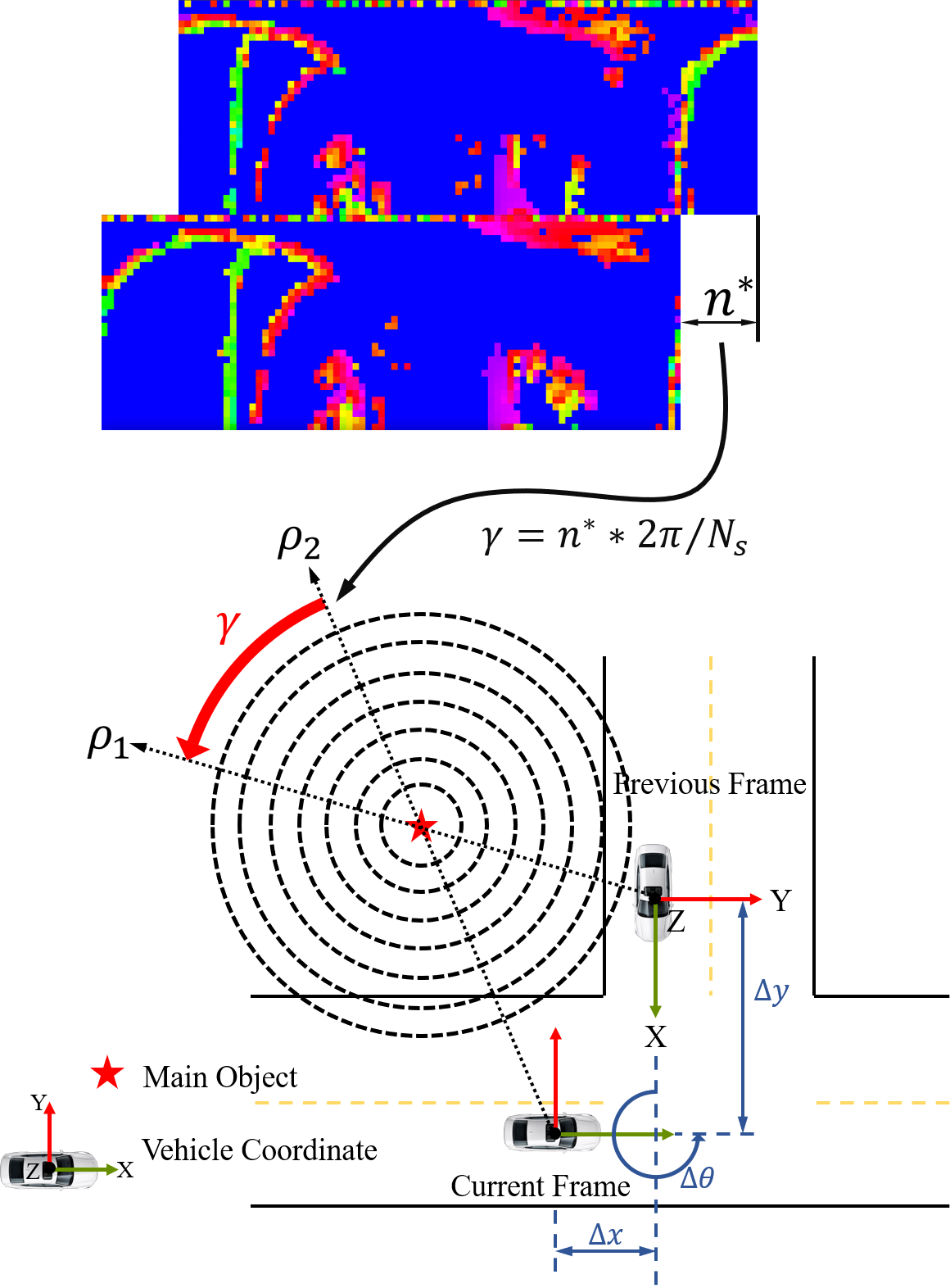}
        
        \caption{An example of contactless loop closure. The relative pose $(\Delta x, \Delta y, \Delta \theta)$ between the current frame and previous frame can be determined by the rotation angle $\gamma$ produced by the matching result $n$ of Object Scan Context and the observation position $\{(x_1, y_1), (x_2, y_2)\}$ of the Main Object in two frames.}
        
        \label{F5}
\end{figure}

\begin{figure*}[htbp]
    \centering
    \subfigure[00]{\includegraphics[width=0.3\textwidth]{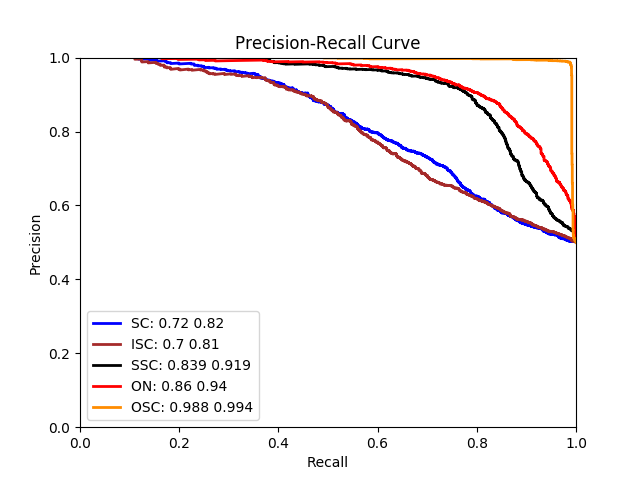}}
    \subfigure[02]{\includegraphics[width=0.3\textwidth]{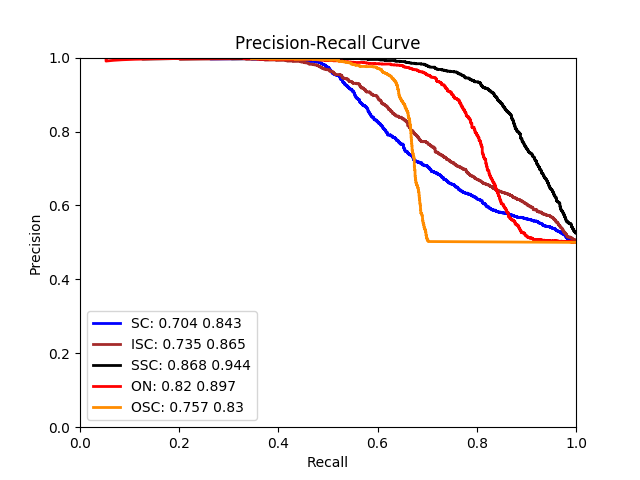}}
    \subfigure[05]{\includegraphics[width=0.3\textwidth]{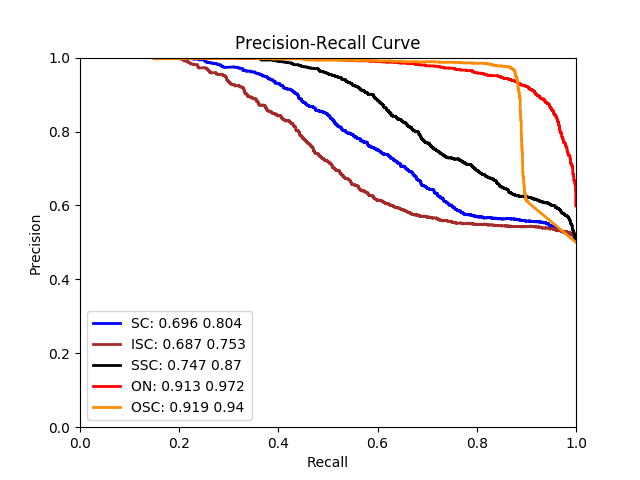}}
    \subfigure[06]{\includegraphics[width=0.3\textwidth]{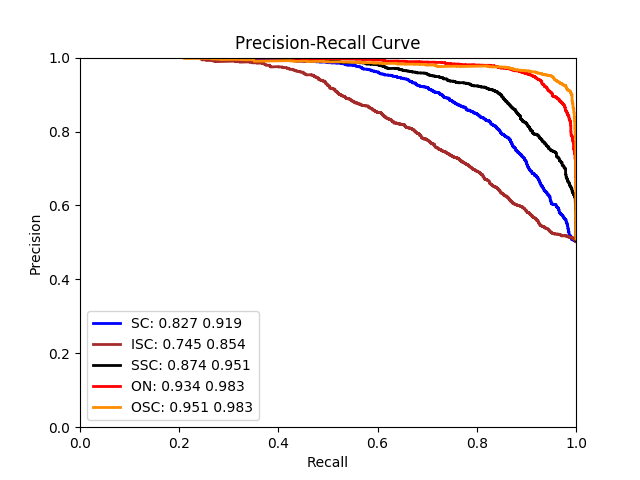}}
    \subfigure[07]{\includegraphics[width=0.3\textwidth]{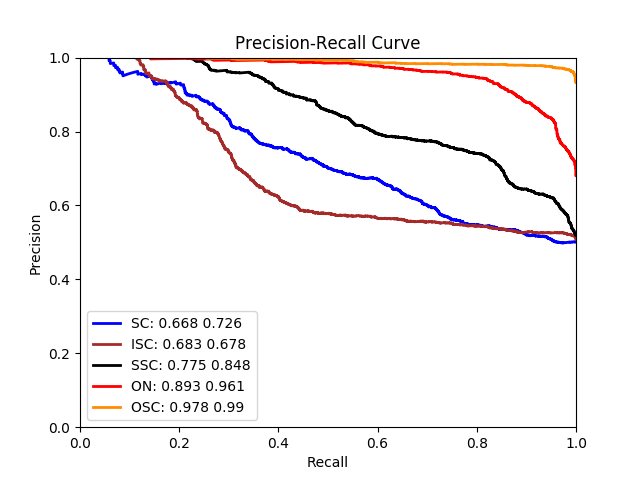}}
    \subfigure[08]{\includegraphics[width=0.3\textwidth]{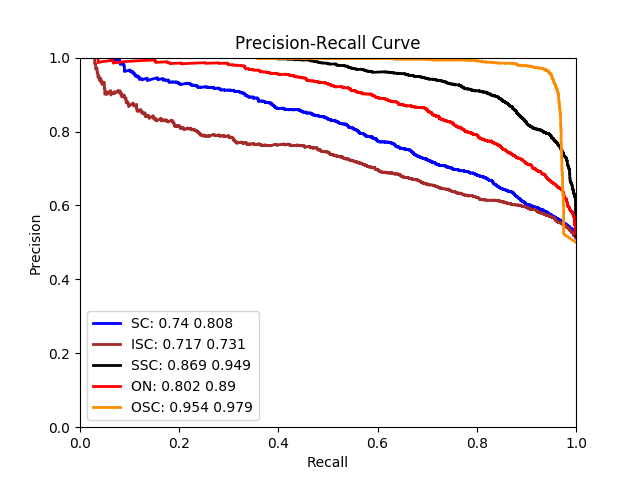}}
    \caption{Precision-Recall curves on KITTI Odometry}
    \label{F6}
\end{figure*}

\subsection{Relative Pose} \label{Relative Pose}
\textbf{Calculate Relative Pose.} \label{Calculate Relative Pose} Existing place recognition methods based on Scan Context rely on ICP or neural networks to calculate the relative pose. ICP with an inaccurate yaw angle as an initial value, is time-consuming and esaily leads to matching errors. Moreover, the neural network requires a large amount of training data and lacks interpretability. However, the Object Scan Context does not rely on the lidar position, allowing for the calculation of the accurate 3-DOF relative pose $(\Delta x, \Delta y, \Delta \theta)$ using an elegant geometric derivation. This calculation requires only the observation position of the Main Object in the two point clouds and the previous offset $n$ in the previous step.

As shown in Fig. \ref{F5}, assuming that the observation position of the Main Object in the candidate frame is $(x_1, y_1)$, and it in the current frame is $(x_2,y_2)$, the relative pose $(\Delta x, \Delta y, \Delta \theta)$ between them can be deduced as:
\begin{equation}
\begin{aligned}
        \Delta x &= x_1-x_2 \cos \Delta \theta + y_2 \sin \Delta \theta \\
        \Delta y &= y_1-x_2 \sin \Delta \theta - y_2 \cos \Delta \theta \\
        \Delta \theta &= \arctan {y_1 \over x_1} - \gamma - \arctan {y_2 \over x_2}
\end{aligned}
\end{equation}
where $\gamma$ indicates the angle corresponding to the precise offset $n^*$ in the previous step:
\begin{equation}
        \gamma = 2\pi \times {n^* \over N_s}
\end{equation}

\textbf{Pose Filter.} \label{Pose Filter} The bubble filter introduced many false matches, thus we proposed an algorithm to eliminate them after calculating the relative poses. These erroneous matching results may happen to be accidentally close to the ground truth in one or two dimensions, but the probability of them being close in all 3-DOF is almost negligible. This phenomenon is referred to as the sparsity of 3-DOF poses.

We utilize this characteristic and project the 3-DOF relative poses $(\Delta x, \Delta y, \Delta \theta)$ onto Cartesian coordinates, with $\Delta x$ and $\Delta y$ represented by metric values and $\Delta \theta$ represented by its sine value. This means that $(5.67, 3.45, 1^\circ)$ and $(5.12, 3.56, 359^\circ)$ are considered to be close, which can significantly aid in calculating distances later on. Then, we perform Euclidean clustering and select the cluster with the largest amount of data to calculate the average value as the result. Finally, the relative pose is provided for point cloud registration, such as ICP.

\begin{table*}[!ht]
    \caption{$F_1$ max scores and Average Precision on KITTI dataset}
    \centering
    \begin{threeparttable}
        \begin{tabular}{cccccccc}
            \toprule
                Methods & 00 & 02 & 05 & 06 & 07 & 08 & Average \\
            \midrule
                SC \cite{SC} & 0.720/0.820 & 0.704/0.843 & 0.696/0.804 & 0.827/0.919 & 0.668/0.726 & 0.740/0.808 & 0.726/0.820 \\

                ISC \cite{ISC} & 0.700/0.810 & 0.735/0.865 & 0.687/0.753 & 0.745/0.854 & 0.683/0.678 & 0.717/0.731 & 0.711/0.782 \\

                SSC \cite{SSC} & 0.839/0.919 & \textbf{0.868}/\textbf{0.944} & 0.747/0.870 & 0.874/0.951 & 0.775/0.848 & 0.869/0.949 & 0.829/0.914 \\

                ON \cite{OVERLAPNET} & 0.860/0.940 & 0.820/0.897 & 0.913/\textbf{0.972} & 0.934/0.983 & 0.893/0.961 & 0.802/0.890 & 0.870/0.941 \\

                OSC & \textbf{0.988}/\textbf{0.994} & 0.757/0.830 & \textbf{0.919}/0.940 & \textbf{0.951}/\textbf{0.983} & \textbf{0.978}/\textbf{0.990} & \textbf{0.954}/\textbf{0.979} & \textbf{0.925}/\textbf{0.953} \\

            \bottomrule
        \end{tabular}
        
        \centering
        \begin{tablenotes}
            \item[*] $F_1$ max scores and Average Precision: $F_1$ max scores / Average Precision.
            \item[**] The best scores are marked in bold.
        \end{tablenotes}
    \end{threeparttable}
        \label{T1}
\end{table*}

\begin{figure*}[htbp]
    \centering
    \subfigure[00]{\includegraphics[width=0.3\textwidth]{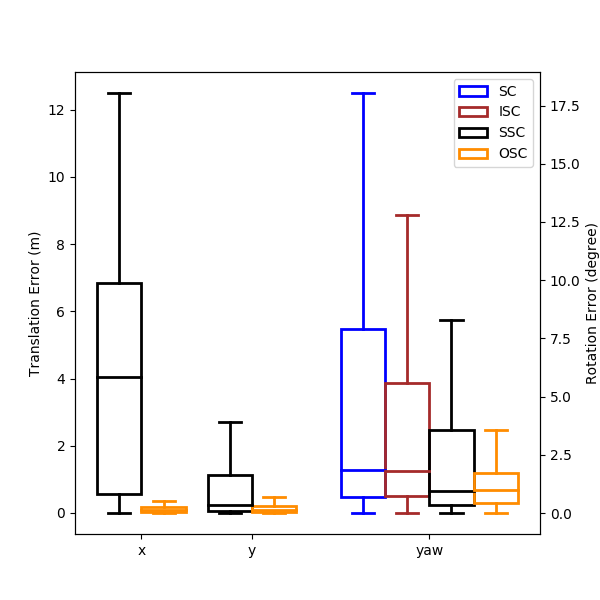}}
    \subfigure[02]{\includegraphics[width=0.3\textwidth]{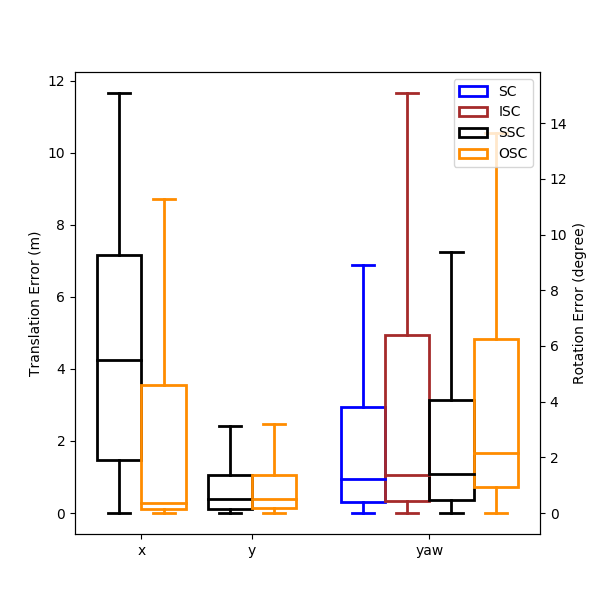}}
    \subfigure[05]{\includegraphics[width=0.3\textwidth]{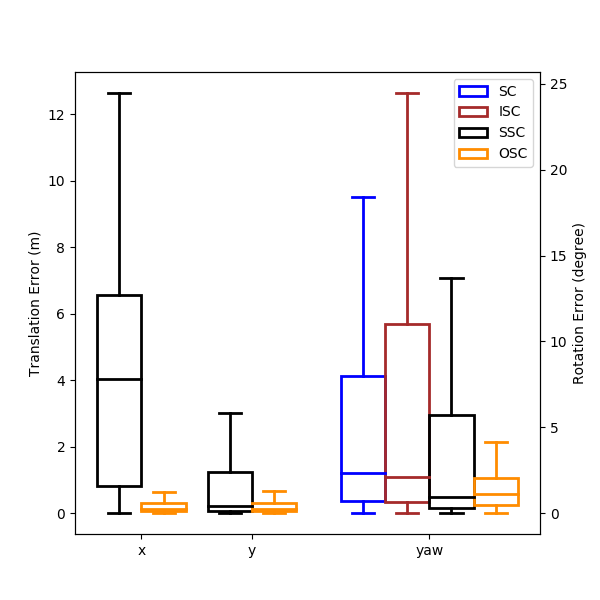}}
    \subfigure[06]{\includegraphics[width=0.3\textwidth]{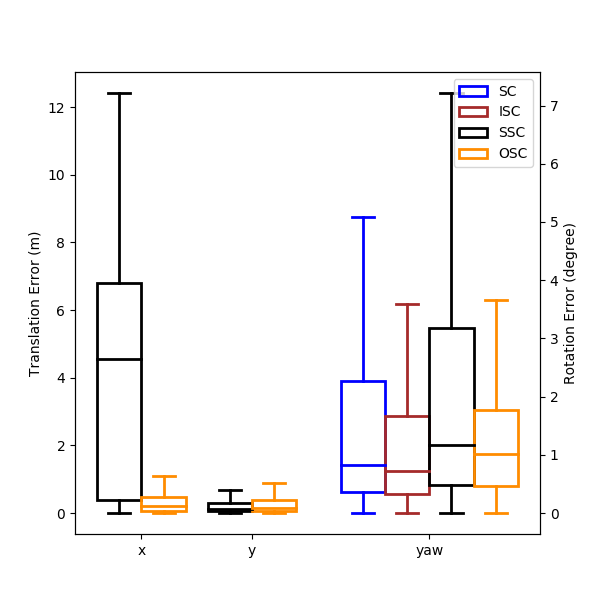}}
    \subfigure[07]{\includegraphics[width=0.3\textwidth]{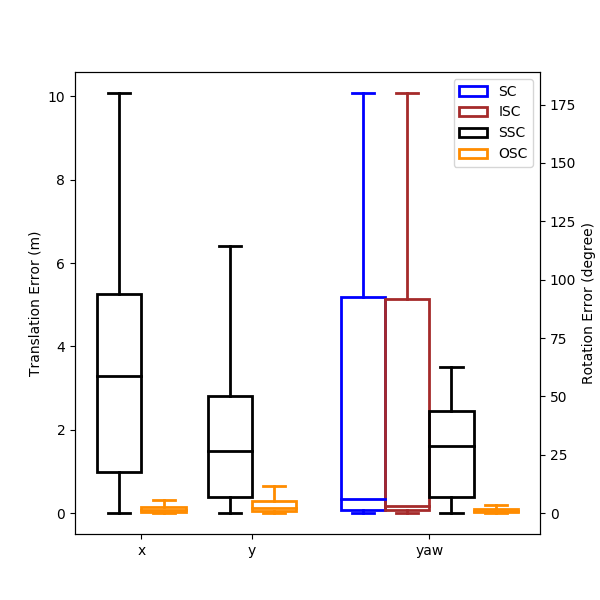}}
    \subfigure[08]{\includegraphics[width=0.3\textwidth]{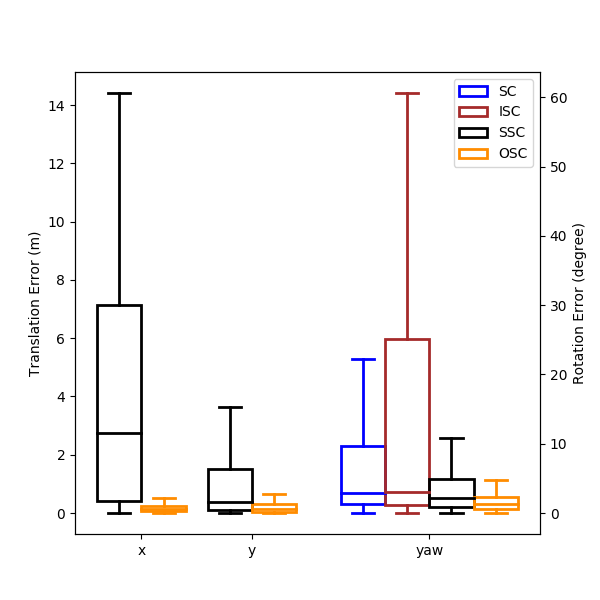}}
    \caption{Translation error and rotation error on KITTI Odometry}
    \label{F7}
    \end{figure*}

\section{EXPERIMENTS} \label{EXPERIMENTS}

\subsection{Dataset and Implementation Details}

We evaluate the performance of our method on KITTI Odometry \cite{KITTI} sequences (00, 02, 05, 06, 07, 08) and KITTI360 \cite{KITTI360} sequence (0009). Prior methods such as SGPR \cite{SGPR} and SSC \cite{SSC} consider two frames of point clouds with a distance less than 3m (greater than 20m) in the real world as positive (negative) samples in performance evaluation, because their descriptors are all egocentric and cannot handle contactless place recognition. However, our method can overcome this limitation, allowing us to use only one threshold (10m in our experiments) to distinguish positive and negative samples, what means that we pay more attention to the detection of various algorithms in contactless place recognition and the accuracy of relative pose. Please note that we tested all other algorithms under the same standard (10m as the standard for dividing positive and negative samples), and all the experimental results below follow this standard. Due to the large number of negative samples (millions) and positive samples (tens of thousands), we randomly select 2,000 positive and 2,000 negative samples from each sequence for our evaluation.

For experiments on the KITTI Odometry we use labels from SemanticKITTI \cite{SEMANTICKITTI}, and for experiments on KITTI360 we use RangeNet++ \cite{RANGENET++} and its pre-trained model to inference. In our experiments, we set $D=20m, N_r=20, N_s=60$ where $D$ denotes the radius of the Object Scan Conetxt, which is local descriptor that only focus on the local environment around the Main Object. All experiments are done on a laptop with AMD Ryzen 7th 5800H.

\subsection{Place Recognition Performance}
We compare our approach with the prior Scan Context methods and the state-of-the-art learning-based methods, including Scan Context \cite{SC} (SC), Intensity Scan Context \cite{ISC} (ISC), Semantic Scan Context \cite{SSC} (SSC), and OverlapNet \cite{OVERLAPNET} (ON). Fig. \ref{F6} shows the precision-recall curve of each method. Tab. \ref{T1} shows the $F_1$ max score and the average precision. The $F_1$ score is defined as:

\begin{equation}
        F_1 = 2 \times {P \times R \over P + R}
\end{equation}

where $P$ and $R$ represent the precision and recall respectively. The average precision (AP) stands for the area enclosed  by the precision-recall curve and the coordinate axes.

In SC and ISC, we set $D = 80m$, $N_r = 20$, $N_s = 60$, where $D$ represents the radius of descriptors. For SSC, we use the ground-truth labels from SemanticKITTI \cite{SEMANTICKITTI}. For OverlapNet, we also build semantic map from the ground-truth labels and use the pre-trained model provided by the author to compute the overlap and yaw.

We can observe from Figure \ref{F6} and Table \ref{T1} that our method achieves almost comprehensive performance superiority over other methods on the [00, 06, 07, 08] sequence, which sufficiently demonstrates the advantages of our method in dealing with contactless place recognition. However, it does not perform well on the [02, 05] sequence, which we attribute to the scarcity of Main Objects in some regions of these two sequences. Moreover, the reason for all curves intersecting at (1, 0.5) is that the number of positive and negative samples selected is the same. It can be interpreted as follows: the algorithm considers all samples to be positive, so the recall rate reaches 1, where the accuracy rate is only affected by the sample distribution (positive samples / all samples).

\begin{figure}[htbp]
        \centering
        \subfigure[0009]{\includegraphics[width=0.45\textwidth]{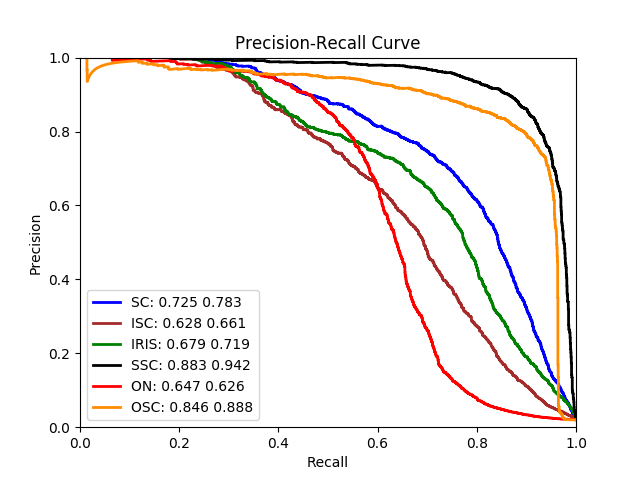}}
        \caption{Precision-Recall curves on KITTI360 (0009)}
        \label{F0}
    \end{figure}

For KITTI360 with more repeated routes, our method can also perform well, as shown in the \ref{F0}. This set of experiments is based on 2000 positive samples and 100000 negative samples, which is why the image is almost intersected at (1, 0).


\begin{table*}[!ht]
        \caption{Translation error and rotation error on KITTI Odometry}
        \centering
        \begin{threeparttable}
        \begin{tabular}{ccccc}
            \toprule
			Seq. & SC & ISC & SSC  & OSC \\
            \midrule
			00      & -/-/1.832 & -/-/1.794 & 4.038/0.224/\textbf{0.943}  & \textbf{0.086}/\textbf{0.103}/0.972 \\
			02      & -/-/\textbf{1.229} & -/-/1.356 & 4.239/0.397/1.383  & \textbf{0.265}/\textbf{0.377}/2.171 \\
			05      & -/-/2.330 & -/-/2.080 & 4.024/0.201/\textbf{0.959}  & \textbf{0.133}/\textbf{0.127}/1.083 \\
			06      & -/-/0.828 & -/-/\textbf{0.727} & 4.564/\textbf{0.116}/1.177  & \textbf{0.206}/0.161/1.010 \\
			07      & -/-/5.978 & -/-/2.816 & 3.279/1.493/28.671 & \textbf{0.067}/\textbf{0.110}/\textbf{0.947} \\
			08      & -/-/2.959 & -/-/2.998 & 2.759/0.368/2.186  & \textbf{0.129}/\textbf{0.131}/\textbf{1.304} \\
			Average & -/-/2.526 & -/-/1.962 & 3.817/0.466/5.886 & \textbf{0.148}/\textbf{0.168}/\textbf{1.248} \\

            \bottomrule
        \end{tabular}
        
        \centering
        \begin{tablenotes}
                \item[*] Translation error and rotation error: $e_x$/$e_y$/$e_{\theta}$. - indicates that the algorithm cannot obtain this result.
		\item[**] The smallest errors are marked in bold.
        \end{tablenotes}
    \end{threeparttable}
        \label{T2}
\end{table*}

\begin{figure}[thpb]
    \centering
    \includegraphics[width=0.45\textwidth]{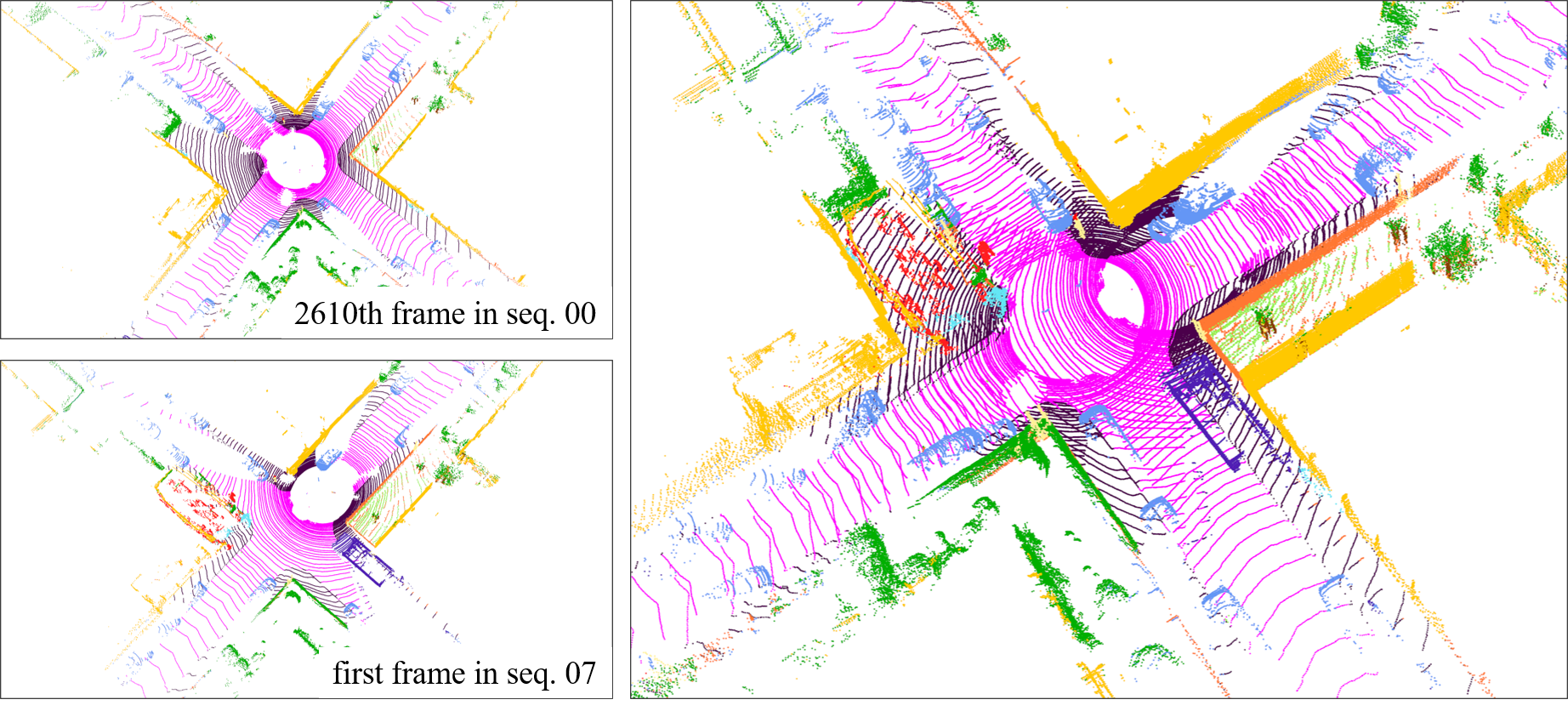}
    
    \caption{An example of robust relocalization performance.}
    
    \label{F8}
\end{figure}

\subsection{Relative Pose Accyracy}
There are not many place recognition mathods mentioned above that can compute the complete relative pose. Among them, SC, ISC and OverlapNet can calculate the yaw angle. SSC can get the 3D pose because of its two-step fast ICP, and our method can calculate the 3D relative pose $(\Delta x, \Delta y, \Delta \theta)$ by the simple coordinate transformation. 

Fig. \ref{F7} shows the relative translation error and rotation error of SSC and OSC. As shown, our method without any point cloud registration algorithm has much smaller error than SSC, which shows our method has advantage in dealing with contactless place recognition. After refined by ICP, this can serve as a global mapping constraint in the SLAM system or the relocalization result. The finer quantization errors are shown in Tab. \ref{T2}, from which we can see that our method outperforms SSC on almost all items.


\subsection{Robust Relocalization Performance}
We evaluated the effectiveness of our method not only for relocalization within the same sequence in KITTI, but also for different sequences. Object Scan Context is centered on the Main Object and represents the local environment around it, allowing accurate point cloud registration even when collected at different times, which is a challenging task for prior methods.

Fig. \ref{F8} demonstrates the relocalization performance between the 2610th frame in KITTI sequence 00 and the first frame in sequence 07. Despite significant changes in the environment, our method still achieves accurate results, which is extremely difficult for prior Scan Context series methods.

\section{CONCLUSIONS}

In this paper, we propose Object Scan Context (OSC), a novel local descriptor for place recognition. Unlike most other lidar-based descriptors, OSC is centered on the Main Object, which enables it to address the problem of contactless place recognition. Moreover, since the descriptor is established independently of the observation position, we can calculate an accurate relative pose through spatial transformation, which is beneficial for SLAM and localization systems.

In future work, we plan to extend the 3-DOF relative pose to 6-DOF and generalize our method to perform well in feature-scarce scenarios such as KITTI Odometry sequence 02.

\addtolength{\textheight}{-12cm}  





\bibliographystyle{IEEEtran}
\bibliography{IEEEexample}

\end{document}